\documentclass[letterpaper, 10pt, conference]{ieeeconf}
\IEEEoverridecommandlockouts 

\usepackage{afterpage}
\usepackage{cite}
\usepackage{algorithm} 
\usepackage{algorithmic}  
\usepackage[algo2e]{algorithm2e} 
\usepackage{amsmath,amssymb,amsfonts}
\usepackage{algorithm}
\usepackage{graphicx}
\usepackage{textcomp}
\def\BibTeX{{\rm B\kern-.05em{\sc i\kern-.025em b}\kern-.08em
    T\kern-.1667em\lower.7ex\hbox{E}\kern-.125emX}}
    
    \usepackage{lineno}
\usepackage{lettrine}
\usepackage{booktabs}
\usepackage{url}
\usepackage{moreverb}

\usepackage{graphicx}
\usepackage{float}
\usepackage{amsmath}
\usepackage{array}
\usepackage{multirow}
\graphicspath{ {Figures/} }

\usepackage{scalerel}
\usepackage{tikz}
\usetikzlibrary{svg.path}

\definecolor{orcidlogocol}{HTML}{A6CE39}
\tikzset{
  orcidlogo/.pic={
    \fill[orcidlogocol] svg{M256,128c0,70.7-57.3,128-128,128C57.3,256,0,198.7,0,128C0,57.3,57.3,0,128,0C198.7,0,256,57.3,256,128z};
    \fill[white] svg{M86.3,186.2H70.9V79.1h15.4v48.4V186.2z}
                 svg{M108.9,79.1h41.6c39.6,0,57,28.3,57,53.6c0,27.5-21.5,53.6-56.8,53.6h-41.8V79.1z M124.3,172.4h24.5c34.9,0,42.9-26.5,42.9-39.7c0-21.5-13.7-39.7-43.7-39.7h-23.7V172.4z}
                 svg{M88.7,56.8c0,5.5-4.5,10.1-10.1,10.1c-5.6,0-10.1-4.6-10.1-10.1c0-5.6,4.5-10.1,10.1-10.1C84.2,46.7,88.7,51.3,88.7,56.8z};
  }
}

\newcommand\orcidicon[1]{\href{https://orcid.org/#1}{\mbox{\scalerel*{
\begin{tikzpicture}[yscale=-1,transform shape]
\pic{orcidlogo};
\end{tikzpicture}
}{|}}}}

\usepackage{hyperref} 

\begin{document}
\title{\LARGE \bf
From Observation to Prediction: LSTM for Vehicle Lane Change Forecasting on Highway On/Off-Ramps}

%

\author{Mohamed Abouras$^{1,2}$ and Catherine~M.~Elias$^{1,2\orcidicon{0000-0002-1444-9816}\,}$,~\IEEEmembership{Member,~IEEE,}%
\thanks{*This work was not supported by any organization}
\thanks{$^{1}$C-DRiVeS Lab: Cognitive Driving Research in Vehicular Systems, Cairo, Egypt
{\tt\small cdrives.researchlab@gmail.com}}%
\thanks{$^{2}$Computer Science and Engineering Department - Faculty of Media Engineering and Technology - German University in Cairo, Egypt}%
\thanks{{\tt\small mohamed.abouras@student.guc.edu.eg, catherine.elias@ieee.org}}%
}

\maketitle
\begin{abstract}
On and off-ramps are understudied road sections even though they introduce a higher level of variation in highway interactions. Predicting vehicles' behavior in these areas can decrease the impact of uncertainty and increase road safety. In this paper, the difference between this Area of Interest (AoI) and a straight highway section is studied. Multi-layered LSTM architecture to train the AoI model with ExiD drone dataset is utilized. In the process, different prediction horizons and different models' workflow are tested. The results show great promise on horizons up to 4 seconds with prediction accuracy starting from about 76\% for the AoI and 94\% for the general highway scenarios on the maximum horizon.
\end{abstract}

\keywords
LSTM, Vehicle Behavior Prediction, Off-ramps, On-ramps, Highway, ExiD, HighD, Deep Learning
\endkeywords

\IEEEpeerreviewmaketitle
\section{Introduction and Related Work}\label{sec1}

\PARstart{D}{riving}  is accompanied by uncertainty; a driver's predictions and reactions are the deciding factors of whether that uncertainty results in an unfortunate outcome, over one million people die from road injuries every year \cite{owid_road_deaths, who_road_traffic_injuries_2023}. Highways add speed to the list of challenges on the road which in turn increases the arbitrary factors present and the outcome lethality. A lot of the classical, physics-based approaches such as constant velocity and constant acceleration models and Kalman filter model which were used in relevant literature have tried to predict the volatile aspects surrounding this dilemma \cite{machineLearningTrajectoryPrediction, enhancedVehicleTrajectory}. These approaches while fast, computationally, they cannot account for complex scene factors like interactions between vehicles \cite{machineLearningTrajectoryPrediction, enhancedVehicleTrajectory}. Alternatively, some classical machine learning (ML) approaches such as support vector machine (SVM) \cite{vehicleGraphEmbeddings} failed to model the hidden state of the entire scene which constrained the capacity for generalization \cite{enhancedVehicleTrajectory}. These models also struggled when the number of input features increased, which made the classification task significantly harder. In addition, manually extracting relevant features can be both challenging and time-consuming. Mistakes made during this early stage of feature annotation can carry through the entire pipeline and negatively affect the model's overall performance\cite{vehicleTrajectoryTopViewDeepLearning}.

Deep Learning (DL) has been a rising tool to overcome classical limitations and has been used to study different scenarios from perceiving the environment to predicting behavior and giving safe decisions. These DL models can construct complex representations of the scene, such as Bird's Eye View (BEV) representations that stack historical perception outputs and HD map information. This capability shows high improvement in generalizing the highly dynamic and varied road environments. While a lot of AI-based research has delved into relevant scenarios, relatively few papers have explored prediction as reviewed by surveys \cite{trajectoryPredictionASurvey, aReviewOfDeepLearningPrediction, surveyOnDeepLearningApproaches, machineLearningTrajectoryPrediction}, which review different DL and ML approaches for vehicle trajectory and motion prediction. Other survey AI techniques for accident prediction and unsafe driving patterns \cite{techniquesForDrivingSafety}. Some studies, as reviewed by \cite{machineLearningTrajectoryPrediction} and \cite{aReviewOfDeepLearningPrediction}, utilize INTERACTION \cite{interactiondataset} dataset which discuss merging lane situations \cite{humanFactorsApproach}. Fewer tried to predict behavior on the highway exiting and merging scenarios. These scenarios are important leverage points on the road as they are nodes of great variation in both behavior and acceleration. Because of the importance of these scenarios, this paper aims to expand on these events in the general context that is the highway. \

Following the methodology of previous literature, similar LSTM networks are utilized to make predictions regarding lane changes at exits and entrances on the highway. This approach is well-supported as LSTMs are a special type of Recurrent Neural Networks (RNNs) that excel at modeling temporal dependencies in sequential data\cite{lstmBasedPredictionAggressive, surveyOnDeepLearningApproaches}. While traditional RNNs mostly struggle with long dependencies, LSTMs overcome this challenge through their selective memory gates.

Prior studies have demonstrated the effectiveness of LSTM-based models for highway lane change and behavior prediction. For instance, \cite{qasemabadiCACC} utilized multi-LSTM trained on HighD dataset and achieved up to 92.43\% accuracy in Cooperative Adaptive Cruise Control (CACC) scenarios by incorporating surrounding vehicles. Others study aggressive lane changes on highways, achieving 98.5\% prediction accuracy \cite{lstmBasedPredictionAggressive}.

Getting accurate lane change predictions can lead to a safer and smoother highway experience. Additionally, the aim of this study is not only to increase the accuracy over the model in the existing literature but also to study key differences between the conditions on the normal highway scenarios and the variations on highway exit and entrance points. This paper makes the following contributions:
\begin{enumerate}
    \item It extends highway behavior studies by specifically focusing on merging and diverging maneuvers at highway entry and exit ramps. 
    \item It compares various multi-layered LSTM architectures to evaluate the effectiveness of end-to-end versus modular learning approaches.
    \item It presents a survey of safe lane change durations from existing literature to inform the selection of different prediction horizons.
    \item It investigates feature bias across different maneuver classes to understand which inputs most influence model predictions.
\end{enumerate}

In section \ref{sec2discuss}, this paper will begin by discussing the methodology of this study. Then, in section \ref{sec3}, the results of the experiments conducted will be reviewed. Section \ref{sec4} concludes the study and proposes future recommendations.   

\begin{figure}
    \centering
    \includegraphics[width=1\linewidth]{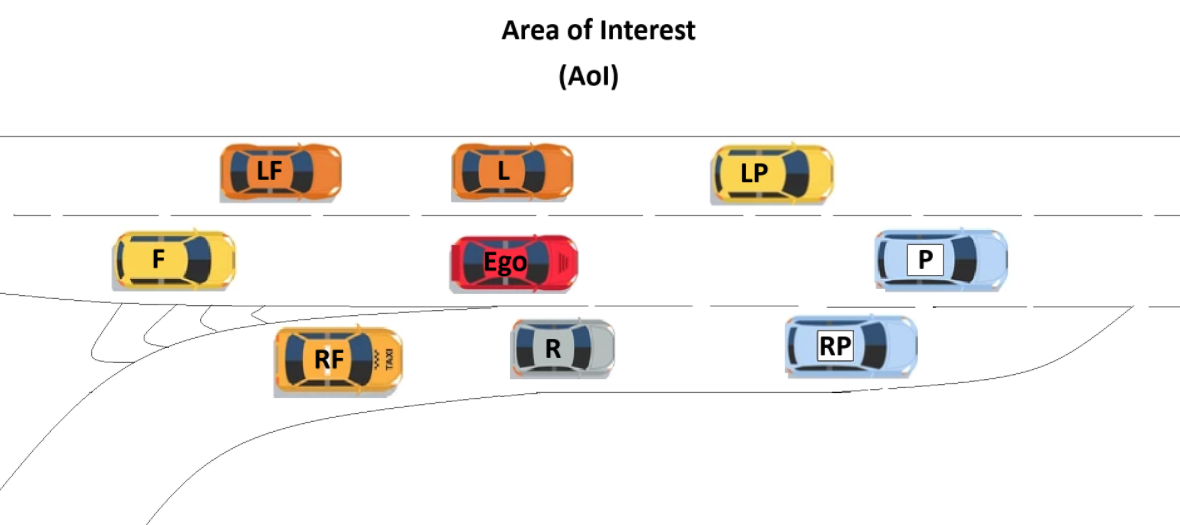}
    \caption{Area of Interest}
    \label{fig:area-of-interest}
\end{figure}

\section{Methodology}\label{sec2discuss}

This section discusses the approach taken to predict lane change maneuvers using a stacked LSTM architecture. First, the ExiD and HighD datasets are prepared and processed. Then, the prediction range's motivation as well as the model architecture and the training process are discussed. \

\subsection{Data Processing and Preparation}\label{subsec21} \

The research environment is defined as structured, where vehicles' speeds are over 100 km/h and traffic rules, such as using designated lanes, are applied.  The Area of Interest (AoI) lies in highway sections that directly connect the highway to the off-ramps (exits) and on-ramps (entrances), as shown in Figure \ref{fig:area-of-interest}.  To study exits and entrances, ExiD drone dataset is used, which is a real and accurate trajectory dataset at highway entries and exits in Germany collected by LevelXData \cite{exiDdataset}. LevelXData also provide another drone trajectory dataset recorded on German highways\cite{highDdataset}, which is used to compare the AoI with general highway scenarios.

\begin{table}[H]
\centering
\caption{Features chosen for training from ExiD and HighD}
\label{tab:dataset-features}

\begin{tabular}{c}\toprule
Dataset Features                   \\\midrule
Ego's Position X(X), Y(Y)\\
Ego's Velocity X(VX), Y(VY)\\
Ego's Acceleration X(AX), Y(AY)\\
Relative Surrounding Distances (dP)\\
Relative Surrounding Velocities (dV)\\
Relative Surrounding Accelerations (dA)\\ \bottomrule \end{tabular}

\end{table}

After the research environment is defined, the data is prepared by recording an observation period of five frames, or one-fifth of a second, for each vehicle (ego) on the road section. This observation period was found to be optimal by Qassemabadi et al. in \cite{qasemabadiCACC}. 
\begin{equation}
dp = \sqrt{(x_s - x_e)^2 + (y_s - y_e)^2}
\label{eq:distance equation}
\end{equation}
\noindent
where $dp$ is the difference in position, with $s$ denoting the surrounding vehicle and $e$ denoting the ego vehicle.
\begin{equation}
dv = \sqrt{(v_s - v_e)^2 + (v_s - v_e)^2}
\label{eq:velocity equation}
\end{equation}
\noindent
where $dv$ is the difference in velocity, with $s$ denoting the surrounding vehicle and $e$ denoting the ego vehicle.
\begin{equation}
da = \sqrt{(a_s - a_e)^2 + (a_s - a_e)^2}
\label{eq:acceleration equation}
\end{equation}
\noindent
where $da$ is the difference in acceleration, with $s$ denoting the surrounding vehicle and $e$ denoting the ego vehicle.

The observed features are ego's position (x, y), velocity (x, y), and acceleration (x, y), found in Table \ref{tab:dataset-features}. The euclidean relative distance eq (\ref{eq:distance equation}) ,  velocity eq (\ref{eq:velocity equation}), and acceleration eq (\ref{eq:acceleration equation}) of the ego's surrounding vehicles are calculated to link between the ego and its acting environment. There are three output classes: Lane Change Left (LCL), Lane Keep (LK), and Lane Change Right (LCR). To avoid class-bias, the original number of entries of each class, found in Table \ref{tab:class-distrib}, was balanced for training and testing. According to the observation period, the models predict the output class after a specified temporal horizon. 

\begin{table}[H]
    \centering
    \caption{Different classes distributions for each scenario after preparation and before balancing}
    \resizebox{\columnwidth}{!}{%
    \begin{tabular}{lclll}\toprule
          &Time in Seconds&  Left&  Keep& Right\\\midrule
          &1&  150645&  36080998& 178715\\
          HighD&2&  271444&  33058674& 320126\\
          &3&  677843&  11174034& 479914\\
          &4&  436755&  27194418& 514418\\
 \multicolumn{5}{c}{-----------}\\
          &1&  316972&  13857584& 200593\\
          ExiD&2&  521974&  3537031& 3537031\\
          &3&  793145&  575544& 575544\\
          &4&  1282296&  16941285& 930741\\ \bottomrule 
    \end{tabular}%
    }
    
    \label{tab:class-distrib}
\end{table}

\subsection{Prediction Horizon}\label{subsec22} \

To predict any action, a reliable reference must be established based on the action's behavior. Thus, to understand lane change maneuvers, an experiment was performed involving two vehicle types--a sedan and an SUV--each driven by experienced drivers on controlled, empty inter-city highways. The drivers exceeded 130 km/h speeds and covered both straight and curved sections, performing 20 lane change maneuvers each. A passenger used a stopwatch to time a full lane change (until the vehicle crosses the lane mark completely) and answered a short survey about the safety perception of the maneuver. Results revealed durations ranging from 2 to 6 seconds. 

\subsection{Model Architecture}\label{subsec23} \


\begin{figure}[H]
    \centering
    \includegraphics[width=1\linewidth]{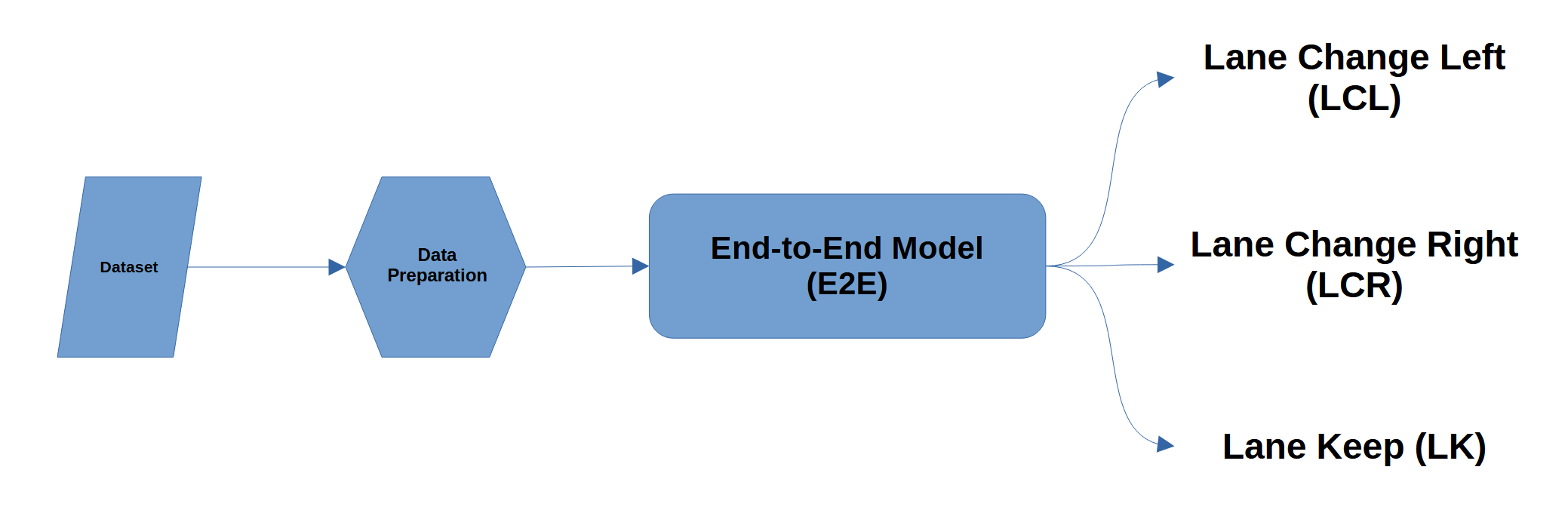}
    \caption{End-to-End Model}
    \label{fig:e2e}
\end{figure}
\begin{figure}[H]
    \centering
    \includegraphics[width=1\linewidth]{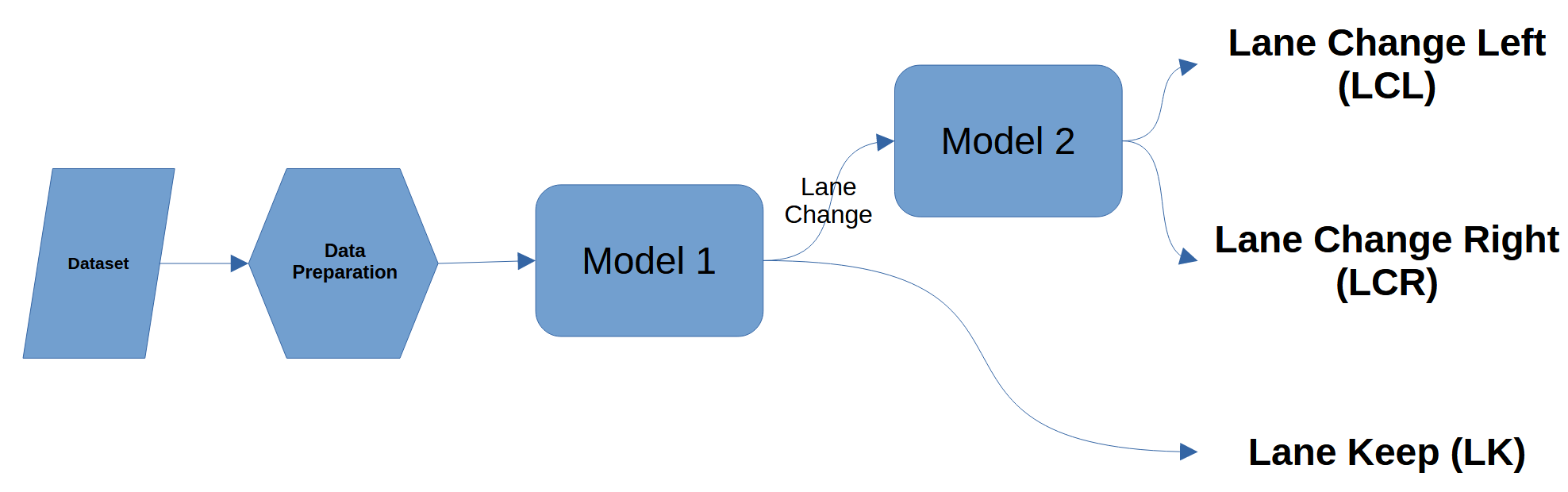}
    \caption{Multi-L Models}
    \label{fig:multiL}
\end{figure}

As the number of learning features for the model is substantial, a stack of multiple LSTMs is used; additionally, as the dataset is relatively huge, overfitting must be avoided. Using multiple LSTMs allows the model to extract hierarchical features and increase long-term memory dependency. It also increases the model's capacity to learn from the given data. Cross-Entropy is used as the loss function, as it directly addresses the classification task. Additionally, a RelU layer is introduced to the model to help prevent convergence issues and learn more complex features. To avoid over-fitting, a dropout rate of 0.2 between the two LSTM layers is used. The model's parameters are set based on the literature review, especially \cite{qasemabadiCACC} on normal highway scenarios. The aforementioned parameters are as shown in Table \ref{tab:model-params}.

There are two models: one model is End-to-End (E2E), shown in Figure \ref{fig:e2e}, where the output is directly defined as either LCL, LCR, or LK; and the other is Multi-Layered (Multi-L), where the first model determines the output as either Lane Change or LK, and if a lane change is detected, the direction is then defined as either Right or Left by the second model, as shown in Figure \ref{fig:multiL}. If the initial Multi-L model can deliver faster binary lane change predictions with less resources, the road user is expected to detect significant road changes in time, making a slower second-staged direction prediction acceptable.

\begin{table}
    \centering
\caption{Model Parameters}
\label{tab:model-params}
    \resizebox{\columnwidth}{!}{%
    \begin{tabular}{ccc}\toprule
         Parameter&  Description& Value\\\midrule
         &  8 surrounding cars, 
5 frames,
(dp, dv, da)& 120\\
         Input Dimension&  5 frames, ego position, 
ego velocity, 
ego acceleration& 30\\
         &  & \\
         Output Dimension&  LCL, LK, LCR for E2E /
(LC, LK), (LCL, LCR)
for Multi-L& 3/1\\
         Batch Size&  Number of training cases per iteration& 32\\
         Hidden Layer Number&  Number of LSTM layers& 2\\
         Dropout Rate&  Percentage of dropout neurons
per iteration& 0.2\\
         Number of Epochs&  Number of training passes
through the dataset& 100\\
         Loss Function&  Function to calculate loss& Cross Entropy\\
         Activation Function&  Function to activate output neurons& RelU\\ 
 Optimizer& Function to minimize prediction errors&RMSProp\\ \bottomrule
    \end{tabular}
}
\end{table}

\subsection{Training Process}\label{subsec24} \


The data is split into a training-to-testing ratio of 80:20. While training, the features affecting class bias are recorded and observed. All models tested are compared for a comprehensive study. 
\section{Results}\label{sec3}

This section explores the results of the experiments. Firstly, the results of the prediction horizon experiment are recorded. Then, the different stacked LSTM models are evaluated. Finally, the feature bias observation is reviewed.


\subsection{Prediction Horizon}
\begin{figure}
    \centering
    \caption{Survey Outcome}
    \includegraphics[width=1\linewidth]{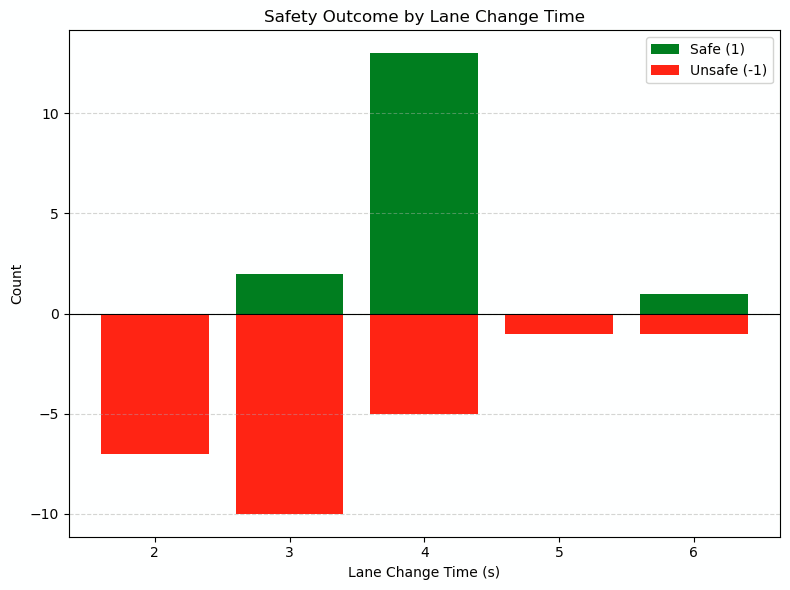}
    
    \label{fig:survey-outcome}
\end{figure}

As the experienced passengers comment on the maneuver, they consider non-empty environments as well in their answers, as per the given follow-up questions. The safe records primarily concentrate on four seconds, and mostly conclude any maneuver under four seconds as very dangerous. There are only three maneuvers from the survey that surpassed four seconds, and they are perceived, mostly, as of unsafe nature, check Figure \ref{fig:survey-outcome}.

\subsection{Model Evaluation}
The models are evaluated by recording the confusion matrix--for detailed breakdown of the predictions--along with the classification report, f1-score, accuracy, precision, recall, number of parameters, model size, inference time, and RAM and GPU allocation. The real prediction validity is evaluated by capturing the real data and comparing it to the prediction output. Predicting lane change direction yields lower prediction accuracy than simply predicting a lane change. Increasing the prediction horizon yields less accurate results in both models. Additionally, the AoI is observed to have added complexity to the models' prediction accuracy. Nevertheless, the models, whether E2E or Multi-L, are flourishing and promising. Check tables \ref{tab:highD-end-to-end}, \ref{tab:highD-model-1}, \ref{tab:highD-model-2}, \ref{tab:AoI-end-to-end}, \ref{tab:AoI-model-1}, and \ref{tab:AoI-model-2} for detailed output.\newline

\begin{table}[H]
\centering
\caption{Training on HighD data only, E2E model}
\label{tab:highD-end-to-end}
\resizebox{\columnwidth}{!}{%
\begin{tabular}{|c|c|c|c|c|}\hline
\cline{2-5}
End-to-End       & \multicolumn{4}{|c|}{Normal Highway Section} \\ \hline
\multicolumn{1}{|c|}{Horizon(seconds)} & \multicolumn{1}{|c|}{1} & \multicolumn{1}{|c|}{2} & \multicolumn{1}{|c|}{3} & \multicolumn{1}{|c|}{4} \\ \hline
Accuracy         & 0.9778    & 0.9844   & 0.9742   & 0.9459   \\ \hline
\multicolumn{1}{|c|}{Precision}        & 0.9737    & 0.9824   & 0.9749   & 0.9485   \\ \hline
Recall           & 0.9824    & 0.9865   & 0.9733   & 0.9430   \\ \hline
\multicolumn{1}{|c|}{F1-Score}         & 0.9779    & 0.9844   & 0.9741   & 0.9457  
\\ \hline\end{tabular}%
}
\end{table}

\begin{table}[H]
\centering
\caption{Training on HighD data only, Model 1: Binary model for lane change.}
\label{tab:highD-model-1}
\resizebox{\columnwidth}{!}{%
\begin{tabular}{|c|c|c|c|c|}\hline
\cline{2-5}
Model 1          & \multicolumn{4}{|c|}{Normal Highway Section}                                                    \\ \hline
\multicolumn{1}{|c|}{Horizon(seconds)} & \multicolumn{1}{|c|}{1} & \multicolumn{1}{|c|}{2} & \multicolumn{1}{|c|}{3} & \multicolumn{1}{|c|}{4} \\ \hline
Accuracy         & 0.9858                & 0.9884                & 0.9763                & 0.9436                \\\hline
\multicolumn{1}{|c|}{Precision}        & 0.9774                & 0.9831                & 0.9755                & 0.9558                \\ \hline
Recall           & 0.9947                & 0.9939                & 0.9772                & 0.9303                \\ \hline
\multicolumn{1}{|c|}{F1-Score}         & 0.9860                & 0.9885                & 0.9764                & 0.9429             
\\ \hline\end{tabular}%
}

\end{table}

\begin{table}[H]
\centering
\caption{Training on HighD data only, Model 2: Binary model for lane change direction.}
\label{tab:highD-model-2}
\resizebox{\columnwidth}{!}{%
\begin{tabular}{|c|c|c|c|c|}\hline
\cline{2-5}
Model 2          & \multicolumn{4}{|c|}{Normal Highway Section} \\ \hline
\multicolumn{1}{|c|}{Horizon(seconds)} & \multicolumn{1}{|c|}{1} & \multicolumn{1}{|c|}{2} & \multicolumn{1}{|c|}{3} & \multicolumn{1}{|c|}{4} \\ \hline
Accuracy         & 1     & 0.9999     & 0.9996    & 0.9951    \\ \hline
\multicolumn{1}{|c|}{Precision}        & 1     & 0.9999     & 0.9998    & 0.995     \\ \hline
Recall           & 1     & 1          & 0.9993    & 0.9941    \\ \hline
\multicolumn{1}{|c|}{F1-Score}         & 1     & 0.9999     & 0.9995    & 0.9945   
\\ \hline\end{tabular}%
}

\end{table}

\begin{table}[H]
\centering
\caption{Training on AoI, E2E model.}
\label{tab:AoI-end-to-end}
\resizebox{\columnwidth}{!}{%
\begin{tabular}{|c|c|c|c|c|}\hline
\cline{2-5}
End-to-End       & \multicolumn{4}{|c|}{Area of Interest}                                                          \\ \hline
\multicolumn{1}{|c|}{Horizon(seconds)} & \multicolumn{1}{|c|}{1} & \multicolumn{1}{|c|}{2} & \multicolumn{1}{|c|}{3} & \multicolumn{1}{|c|}{4} \\ \hline
Accuracy         & 0.8279                & 0.8229                & 0.7809                & 0.7688                \\ \hline
\multicolumn{1}{|c|}{Precision}        & 0.8204                & 0.8174                & 0.7666                & 0.7534                \\ \hline
Recall           & 0.8289                & 0.8092                & 0.7497                & 0.7287                \\ \hline
\multicolumn{1}{|c|}{F1-Score}         & 0.8244                & 0.8128                & 0.7569                & 0.7376            
\\ \hline\end{tabular}%
}

\end{table}

\begin{table}[H]
\centering
\caption{Training on AoI, Model 1: Binary model for lane change.}
\label{tab:AoI-model-1}
\resizebox{\columnwidth}{!}{%
\begin{tabular}{|c|c|c|c|c|}\hline
\cline{2-5}
Model 1          & \multicolumn{4}{|c|}{Area of Interest}                                                          \\ \hline
\multicolumn{1}{|c|}{Horizon(seconds)} & \multicolumn{1}{|c|}{1} & \multicolumn{1}{|c|}{2} & \multicolumn{1}{|c|}{3} & \multicolumn{1}{|c|}{4} \\ \hline
Accuracy         & 0.8777                & 0.8749                & 0.8547                & 0.8331                \\ \hline
\multicolumn{1}{|c|}{Precision}        & 0.8387                & 0.8768                & 0.8595                & 0.8390                \\ \hline
Recall           & 0.9070                & 0.8736                & 0.8595                & 0.8250                \\ \hline
\multicolumn{1}{|c|}{F1-Score}        & 0.8715                & 0.8752                & 0.8542                & 0.8319              
\\ \hline\end{tabular}%
}

\end{table}

\begin{table}[H]
\centering
\caption{Training on AoI, Model 2: Binary model for lane change direction.}
\label{tab:AoI-model-2}
\resizebox{\columnwidth}{!}{%
\begin{tabular}{|c|c|c|c|c|}\hline
\cline{2-5}
Model 2          & \multicolumn{4}{|c|}{Area of Interest} \\ \hline
\multicolumn{1}{|c|}{Horizon(seconds)} & \multicolumn{1}{|c|}{1} & \multicolumn{1}{|c|}{2} & \multicolumn{1}{|c|}{3} & \multicolumn{1}{|c|}{4} \\ \hline
Accuracy         & 0.9669  & 0.9454  & 0.934   & 0.8216 \\ \hline
\multicolumn{1}{|c|}{Precision}        & 0.9665  & 0.9352  & 0.9199  & 0.8299 \\ \hline
Recall           & 0.9762  & 0.9721  & 0.9742  & 0.8566 \\ \hline
\multicolumn{1}{|c|}{F1-Score}         & 0.9713  & 0.9533  & 0.9462  & 0.843 
\\ \hline\end{tabular}%
}

\end{table}

The following points can be concluded from the results in the tables: \newline
\begin{enumerate}
    \item Even the lowest prediction accuracy for a 4-second horizon is quite promising.
    \item Since E2E and Multi-L models have very similar accuracies, the accuracy will not be the determinant factor for which model to adopt.
\end{enumerate}
According to the rest of the evaluations, found in Table \ref{tab:models-multi-data} and Table \ref{tab:models-compare}, the following is concluded: \newline
\begin{enumerate}
    \item All three models (E2E, Multi-L model 1, and Multi-L model 2) have similar size, inference time, and number of parameters. 
    \item RAM and GPU allocated for Multi-L model 1 is the greatest, even greater than the E2E model, making it the most consuming of system resources. 
    \item Using the Multi-L approach will roughly double all resources and time compared to using the E2E approach.
\end{enumerate}

\begin{table}[H]
\centering
\caption{Data of number of parameters, size, RAM and GPU used, and inference time for the multi-layer models.}
\label{tab:models-multi-data}
\resizebox{\columnwidth}{!}{%
\begin{tabular}{|c|c|c|}
\hline
\multirow{2}{*}{} & \multicolumn{2}{c|}{Multi Layer} \\ \cline{2-3}
                  & Model 1 & Model 2 \\ \hline
Number of Parameters & 275,585 & 275,585 \\ \hline
Size & $\sim$1.05MB & $\sim$1.05MB \\ \hline
RAM Used & 3414.17 MB & 2201.97 MB \\ \hline
GPU Used &
  \begin{tabular}[c]{@{}c@{}}Allocated: 495.44 MB\\ Reserved: 762.00 MB\end{tabular} &
  \begin{tabular}[c]{@{}c@{}}Allocated: 240.62 MB\\ Reserved: 552.00 MB\end{tabular} \\ \hline
Inference Time &
  \textbf{t = $\sim$1.5 seconds} &
  \textbf{t = $\sim$1.5 seconds} \\ \hline
\end{tabular}%
}
\end{table}

\begin{table}[H]
\centering
\caption{Comparing Multi-Layer full pipeline to End-to-End model}
\label{tab:models-compare}
\resizebox{\columnwidth}{!}{%
\begin{tabular}{|c|c|c|c|c|}
\hline
\multirow{2}{*}{} & \multicolumn{3}{c|}{Multi-L Models} & E2E Model\\ \cline{1-5}
Number of Parameters & \multicolumn{3}{c|}{551,170} & 275,843 \\ \hline
Size & \multicolumn{3}{c|}{Both models sizes} & $\sim$1.05MB \\ \hline
RAM Used & \multicolumn{3}{c|}{5,616.14 MB} & 3302.57 MB \\ \hline
GPU Used & \multicolumn{3}{c|}{Allocated: 736.06 MB} &
\begin{tabular}[c]{@{}c@{}}Allocated: 285.57 MB \\ Reserved: 550.00 MB\end{tabular} \\ \hline
Inference Time & \multicolumn{3}{c|}{Roughly 2 × t, worst case} & t = $\sim$1.05 seconds \\ \hline
\end{tabular}%
}
\end{table}

\begin{figure}[H]
    \centering
    \includegraphics[width=1\linewidth]{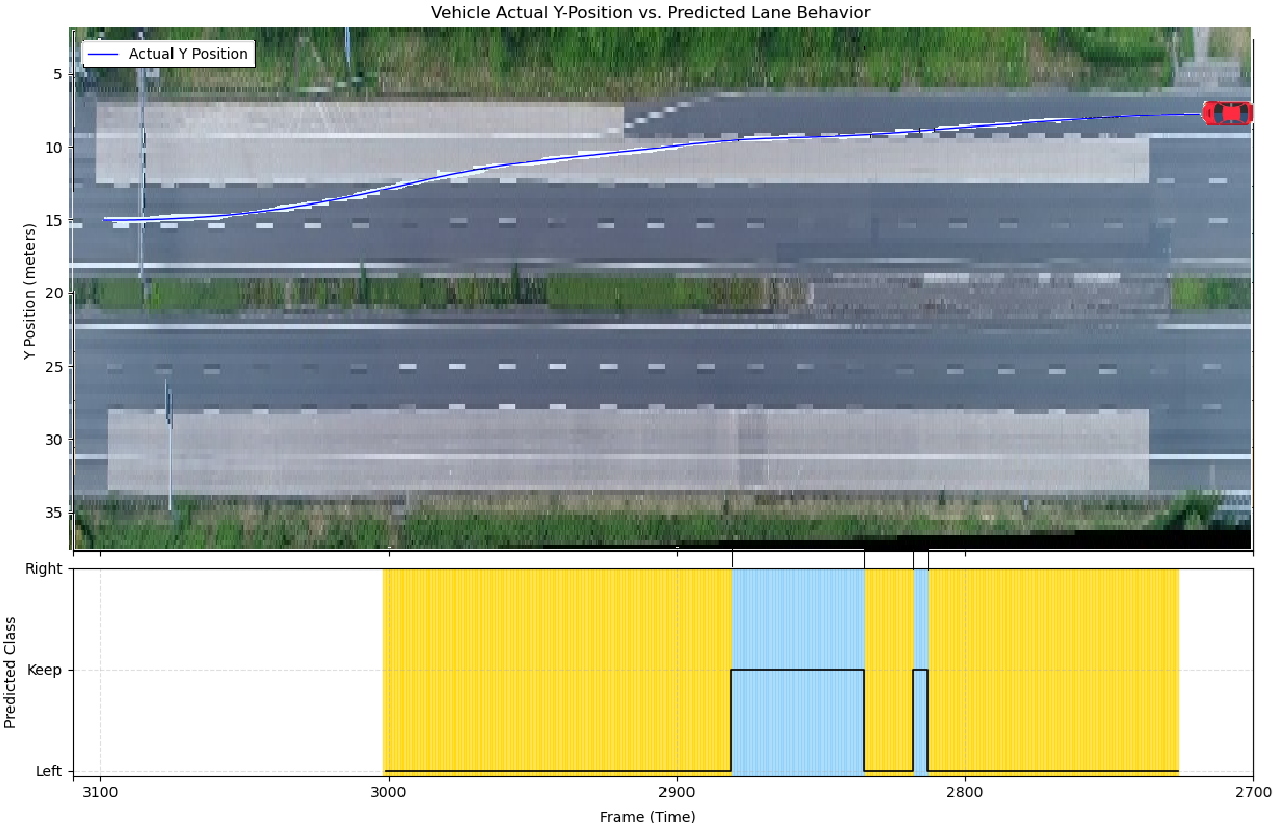}
    \caption{Vehicle path example vs Prediction over time.}
    \label{fig:vehicle-path}
\end{figure}

Accordingly, the best model approach in accuracy, resources, and inference is the E2E model. The model's output was then studied on random cases, such as in Figure \ref{fig:vehicle-path}, and it was found that even when the model fails with the normal evaluation, it correctly evaluates that the vehicle is moving noticeably in the direction of a lane change, even if a lane change doesn't necessarily occur after the prediction horizon. By effectively recognizing subtle motion cues, the model shows strong practical utility and promise for reliable use in real-world scenarios. It is understood, however, that because of the 1-second inference delay, the model with 1-second prediction horizon becomes obsolete and that for other models, inference delay reduces the available reaction time for road users.

\subsection{Feature Bias Observation}
\begin{table}[H]
\centering
\caption{Top Feature Biases per Class: Left (1), Keep (0), Right (2)}
\label{tab:feature-importance}
\begin{tabular}{|c|l|l|l|}
\hline
\textbf{\#} & \textbf{Left (Class 1)} & \textbf{Keep (Class 0)} & \textbf{Right (Class 2)} \\
\hline
1  & t3\_ego\_x& t2\_ego\_x& t0\_ego\_x\\
2  & t0\_ego\_x& t1\_ego\_x& t1\_ego\_x\\
3  & t0\_ego\_y& t4\_ego\_x& t2\_ego\_x\\
4  & t1\_ego\_y& t2\_leftFollowing\_dp& t1\_ego\_y\\
5  & t3\_ego\_y& t4\_rightPreceding\_dp& t4\_ego\_y\\
6  & t2\_ego\_x& t4\_ego\_y& t2\_ego\_y\\
7  & t2\_ego\_y& t4\_leftFollowing\_dp& t3\_ego\_x\\
8  & t2\_leftPreceding\_dp& t1\_leftFollowing\_dp& t2\_leftPreceding\_dp\\
9  & t3\_leftFollowing\_dp& t1\_ego\_y& t2\_rightPreceding\_dp\\
10 & t1\_ego\_x& t3\_ego\_y& t3\_rightPreceding\_dp\\
11 & t4\_leftPreceding\_dp& t0\_ego\_x& t3\_ego\_y\\
12 & t3\_leftPreceding\_dp& t0\_ego\_y& t3\_following\_dp\\
13 & t4\_ego\_y& t3\_rightPreceding\_dp& t4\_leftFollowing\_dp\\
14 & t0\_leftFollowing\_dp& t3\_following\_dp& t3\_leftPreceding\_dp\\
15 & t2\_preceding\_dp& t0\_following\_dp& t0\_ego\_y\\
\hline
\end{tabular}
\end{table}
Since the dataset is originally unbalanced, it is integral to observe which features bias the E2E model to each output class. The report in Table \ref{tab:feature-importance} shows the observed most influential 15 features, ordered from greatest to least influence, on the model according to each class. The conclusion of the report is the following:
\begin{itemize}
\item The features common in influencing the model are: the vehicle's position and the Left Following vehicles' dp. 
    \item The features biasing the model towards Class Left are: Left Preceding and Preceding vehicles' dp.
    \item The features biasing the model towards Class Keep are: Right Preceding and Following vehicles' dp.
    \item The features biasing the model towards Class Right are: Left Preceding, Right Preceding, and Following vehicles' dp.
\end{itemize}
\section{Conclusion and Future Recommendations}\label{sec4}
In conclusion, a non-complex model such as two stacked LSTMs can predict with high accuracy whether a vehicle will keep its lane or change it and in which direction. It is noted that an inference time of one second decreases the decision period. The AoI is found to be a promising domain to expand upon as it poses higher uncertainty than normal highway scenarios. While the prediction horizon experiment found that 4 seconds is a reasonable time to complete a lane change, there were many cases of vehicles taking more than this period in the dataset. The models still hold promise for further study at longer temporal horizons despite their focus on a 1-second to 4-second prediction horizon. One E2E model is found to be better than a Multi-L model in speed and resource consumption. Exploring longer prediction horizons and different architectures is recommended. Additionally, a generative path model that confidently generates future scenarios is proposed for future work. Applying the proposed E2E model can leverage the confident generated paths to estimate potential future maneuvers over extended prediction horizons.


\appendices

\bibliographystyle{IEEEtran}
\bibliography{sections/ref} 

\end{document}